\documentclass[11pt]{article}

\usepackage[final]{acl}

\usepackage{times}
\usepackage{latexsym}

\usepackage[T1]{fontenc}

\usepackage[utf8]{inputenc}

\usepackage{microtype}

\usepackage{inconsolata}

\usepackage{graphicx}
\usepackage{tcolorbox}
\tcbuselibrary{skins}
\usepackage{xcolor}
\usepackage{colortbl}
\definecolor{tabgroup}{gray}{0.90}
\usepackage{amsmath}
\newcommand{\method}{\textsc{Hint}\xspace}
\newcommand{\methodlong}{Hindsight-INformed Turn-level self-distillation}
\newcommand{\taubench}{$\tau$\text{-}\textsc{bench}\xspace}
\newcommand{\taur}{$\tau$\text{-}\textsc{bench}\text{-}\textsc{Retail}\xspace}
\newcommand{\taua}{$\tau$\text{-}\textsc{bench}\text{-}\textsc{Airline}\xspace}
\newcommand{\webshop}{\textsc{WebShop}\xspace}
\newcommand{\demosd}{SD~(demo)\xspace}
\newcommand{\envsd}{SD~(environment)\xspace}
\newcommand{\outsd}{SD~(outcome)\xspace}
\newcommand{\sfthint}{SFT~(hint)\xspace}
\newcommand{\rlhint}{RL~(hint)\xspace}
\newcommand{\G}{G}

\newcommand{\hist}{H_t}
\newcommand{\phist}{\widetilde{H}_t}

\usepackage{xspace}
\usepackage{pifont}
\usepackage{amssymb}
\usepackage{booktabs}
\usepackage{algorithm}
\usepackage{algpseudocode}
\algrenewcommand\algorithmiccomment[1]{\quad{\color{black!55}// #1}}
\makeatletter
\newcommand{\LComment}[1]{\Statex\hskip\ALG@thistlm{\color{black!55}// #1}}
\makeatother
\usetikzlibrary{calc,arrows.meta,backgrounds}
\usepackage{multirow}
\usepackage{enumitem}
\title{Multi-Turn Agent Self-Distillation with Compressed Hindsight}

\author{
  \textbf{Haoran Liu\textsuperscript{1 $\ast$}},
  \textbf{Yuwei Zhang\textsuperscript{1 $\ast$}}, 
  \textbf{Xiyao Li\textsuperscript{2}},
  \textbf{Bohan Lyu\textsuperscript{3}},
  \textbf{Jingbo Shang\textsuperscript{1 $\dagger$}} \\
  \vspace{0.15cm}
  \textsuperscript{1}University of California, San Diego \\
  \textsuperscript{2} Independent Researcher 
  \textsuperscript{3} University of California, Berkeley\\
\\
  \small{
    \textsuperscript{$\ast$}Equal contribution. 
  }
\\
  \small{
    \textsuperscript{$\dagger$}\textbf{Correspondence:} \href{mailto:jshang@ucsd.edu}{jshang@ucsd.edu} 
  }
}

\begin{document}
\maketitle

\begin{abstract}
 In reinforcement learning for multi-turn agents, a sparse
 outcome reward makes credit assignment across intermediate tool calls
 difficult.
On-policy self-distillation eases credit assignment by having the model teach
itself with dense supervision: the model with privileged information (the teacher) scores each token
generated by the model without it (the student).
However, it remains unclear what forms of feedback are most effective in multi-turn agentic settings.
We find that the natural multi-turn variant fails: when the
teacher is given a complete successful demonstration at every turn, training
makes the agent worse than the untrained model. The teacher ends up rewarding
the student for copying the demonstration, even when a copied action is not
right in the student's context. We propose \method, in which a reflector, by default the policy
itself, reads the completed trajectory and writes short turn-level feedback.
The teacher then sees what went wrong at each turn, not the actions that followed.
On \taubench and \webshop, \method matches GRPO with one-eighth as many rollouts.
\end{abstract}

\section{Introduction}
\label{sec:intro}

\begin{figure}[t!]
\centering\footnotesize
\includegraphics[width=\linewidth]{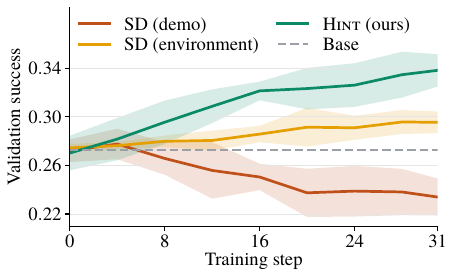}
\caption{\textbf{Validation success of three self-distillation teachers}
on \taur{}.
The teachers differ only in their privileged context:
\demosd{} uses a successful demonstration of the same task;
\envsd{} uses environment feedback at the current turn;
and \method{} (ours) uses a reflected diagnosis from the completed rollout.
Base denotes the untrained model.
Bands show variation.}
\label{fig:curves}
\end{figure}

\begin{figure*}[t]
\centering
\includegraphics[width=\textwidth]{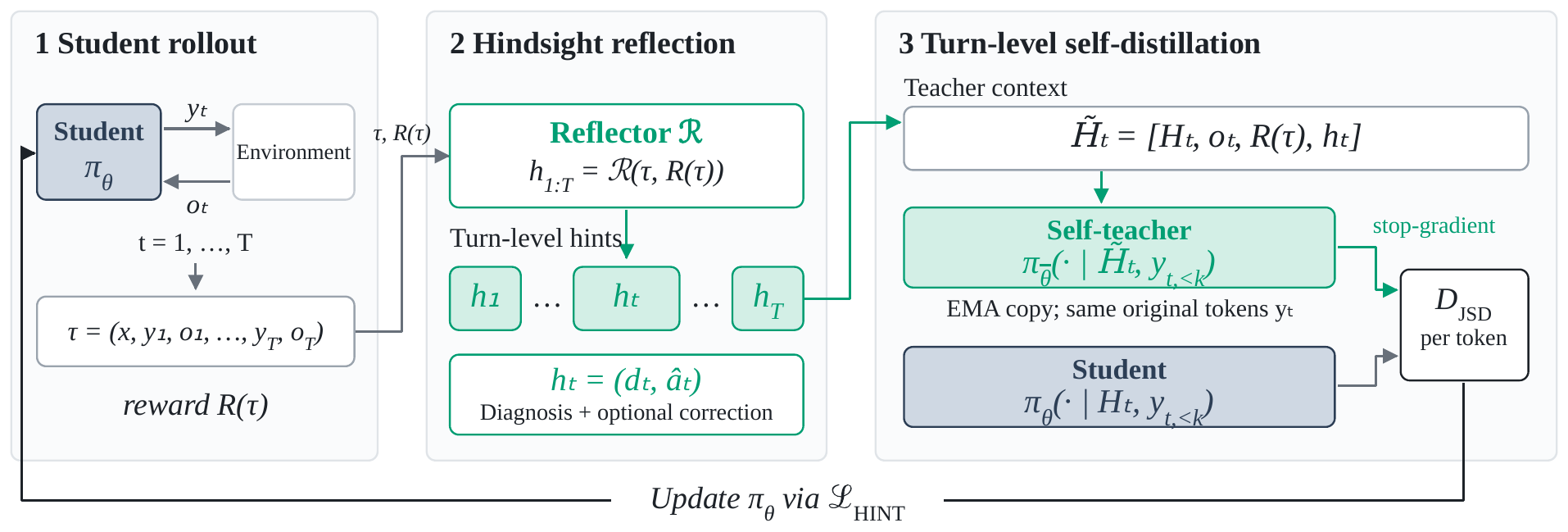}
\caption{\textbf{\method in three stages.} (1) The student rolls out
under the ordinary causal history and the environment returns a terminal
reward. (2) A reflector reads the completed trajectory and its outcome
and compresses them into per-turn diagnoses $h_{1:T}$. (3) For each
scored turn, a stop-gradient self-teacher re-scores the student's
\emph{original} tokens under $\phist$, and the student is trained to match
it.}
\label{fig:pipeline}
\end{figure*}

Large language model agents solve tasks by interacting with tools over many
turns \citep{yao2022react,schick2023toolformer,patil2024gorilla}.
When task outcomes can be scored automatically, reinforcement learning (RL) provides a natural post-training
objective. In many agent environments, however, reward is observed only after
the interaction ends, through a database post-condition, final-state check,
or test suite \citep{yao2024tau,yao2022webshop,trivedi2024appworld}.
A single outcome reward must therefore assign credit across many intermediate
decisions. Failed rollout groups can provide no reward contrast, while
successful trajectories may reinforce unnecessary or incorrect intermediate
actions together with the actions required for success.

On-policy self-distillation provides a different source of supervision.
A self-teacher conditions on privileged information---such as execution feedback, a reference solution, or a hint---and provides token-level targets for outputs sampled from the current policy
\citep{hubotter2026reinforcement,zhao2026self,shenfeld2026selfdistillationenablescontinuallearning}.
Existing self-distillation methods have been developed primarily for single-turn tasks, leaving their extension to multi-turn agents largely unexplored
\citep{zhang2026learning,ye2026policy,penaloza2026privileged}.

\paragraph{The natural multi-turn extension fails.}
We find that directly extending self-distillation to multi-turn agents can be
harmful. The natural extension gives the teacher a successful trajectory
$\tau^{*}$ of the same task, sampled in advance by another model, at every turn
and uses the teacher to re-score each
action in the student's rollout; we call this variant \demosd, for
self-distillation with a demonstration. The feedback is dense and informative, yet
training with it reduces success on \taur from $27.2$ to $22.5$ and on
\webshop from $21.8$ to $17.4$, below the corresponding untrained models
(Figure~\ref{fig:curves} shows the training
curve).

We attribute the failure to what the demonstration lets the teacher reward.
When the student acts at turn $t$, it sees only its own interaction history. The
teacher also sees the demonstration, including what happens in later turns.
As a result, the teacher judges each action by whether it matches the
demonstration, not by whether it was a good decision at turn $t$. An action
that matches the demonstration is not necessarily the best action for the
student.
Multi-turn self-distillation therefore needs
privileged context that keeps the useful hindsight, namely what went wrong at
turn $t$, without later actions to match.

\paragraph{Compressed hindsight.}
We propose \method{} (\methodlong), which converts trajectory-level hindsight into compact turn-level diagnoses. 
After each rollout, a reflector reads the completed trajectory and outcome and
produces a diagnosis $h_t$ for each assistant turn.
The self-teacher then re-scores the student's original action using the student's decision context together with the corresponding turn-level diagnosis, rather than the raw future trajectory.
The diagnosis is used only as privileged context for the teacher and is not directly imitated.

\paragraph{Contributions.}
\begin{itemize}
[leftmargin=*,topsep=2pt,itemsep=2pt,parsep=0pt]
\item We identify a failure mode of multi-turn self-distillation: conditioning
the teacher on a complete successful trajectory, as in \demosd,
can reduce performance below the base model, because the teacher rewards copying
the demonstration rather than the right decision at the turn being scored. We
analyze the resulting supervision at the token level.

\item We propose \method{}, which compresses trajectory-level hindsight into
turn-level privileged feedback before self-distillation, preserving useful
information from later interaction without exposing the teacher to the raw
future trajectory.

\item We compare alternative ways of using the same hindsight signal,
including supervised fine-tuning, scalar turn-level rewards, and token-level
self-distillation. Across three benchmarks and two model scales, \method{}
matches or outperforms GRPO while using one eighth as many environment
rollouts and reducing interaction length.
\end{itemize}

\section{Preliminaries}
\label{sec:prelim}

\begin{figure*}[t]
\centering
\includegraphics[width=\textwidth]{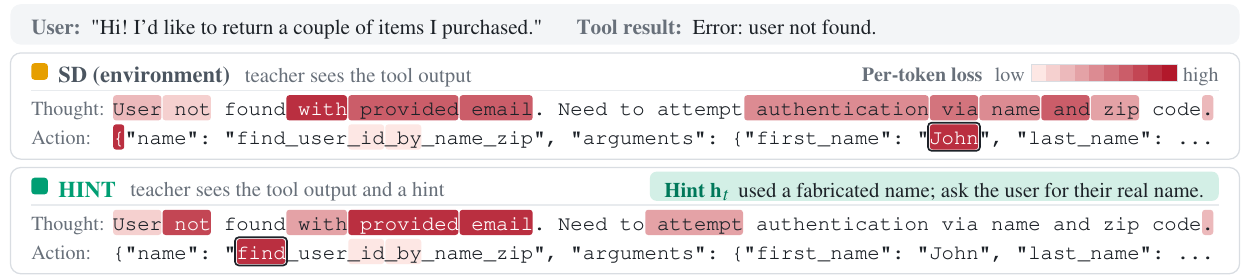}
\caption{\textbf{Compressed hindsight shifts token-level supervision to the decision fork.} The agent fabricates
identity information and calls \texttt{find\_user\_id\_by\_name\_zip}; the
environment replies only \texttt{user not found}. The \envsd teacher
concentrates loss on the placeholder argument---the symptom the observation
named---while \method, given the diagnosis, moves it onto the tool choice at
the decision fork. Table~\ref{tab:ablation} is the same effect aggregated over all
wrong-tool turns.}
\label{fig:tokloss}
\end{figure*}

\paragraph{Multi-turn agent interaction.}
Given a task instruction $x$, the agent interacts with an environment for at
most $T$ turns. At turn $t$ the policy conditions on the interaction history
$\hist = (x, y_1, o_1, \dots, y_{t-1}, o_{t-1})$. $y_i$ is an assistant
action: a reasoning span followed by a tool call.
$o_i$ is the observation the environment returns for it. The student samples
$y_t \sim \pi_\theta(\cdot\mid \hist)$, and a rollout is
$\tau = (x, y_1, o_1, \dots, y_T, o_T)$. The environment emits
terminal reward $R(\tau)$ only at the end of the episode.
We write $\tau_{>t}=(y_{t+1},o_{t+1},\dots,y_T,o_T)$ for the suffix of the
trajectory after turn $t$, and $\G$ for the number of rollouts sampled per
task. Each action is a token sequence
$y_t = (y_{t,1},\dots,y_{t,|y_t|})$, and $y_{t,<k}$ denotes its first $k-1$
tokens. Appendix~\ref{app:notation} collects all symbols.

\paragraph{On-policy self-distillation.}
On-policy distillation (OPD) supervises the student on contexts it generated
itself, which avoids the train--inference mismatch of distilling on a fixed
corpus \citep{agarwal2024policy,lu2025onpolicydistillation}. \emph{Self}-distillation
removes the external teacher: the teacher is the same policy (or a delayed
copy, written $\pi_{\bar\theta}$)
evaluated under a
\emph{privileged context} $c$ that the student did not have
\citep{snell2022learning,hubotter2026reinforcement}. The student is trained to
match the teacher's distribution over its own tokens,
\begin{equation}
\mathcal{L}_{\text{self-OPD}}(\theta) = \mathbb{E}_{\tau\sim\pi_\theta}
\!\left[\sum_{t}\sum_{k} D\!\left(p^{\text{stu}}_{t,k},\, p^{\text{tea}}_{t,k}\right)\right],
\label{eq:selfopd}
\end{equation}
with
$p^{\text{stu}}_{t,k} = \pi_\theta(\cdot\mid \hist, y_{t,<k})$ and
$p^{\text{tea}}_{t,k} = \mathrm{sg}\!\left[\pi_{\bar\theta}(\cdot\mid c, y_{t,<k})\right]$,
where $\mathrm{sg}[\cdot]$ stops gradients on the teacher side. Both distributions are
taken over the \emph{same} student-generated tokens: the privileged context
changes what the teacher knows, never what is being scored.

\paragraph{Jensen--Shannon divergence.}
We instantiate $D$ as the symmetric Jensen--Shannon divergence (JSD), i.e.,
the generalized JSD with $\alpha=0.5$. With
mixture
$m = \tfrac{1}{2}(p+q)$,
\begin{equation}
D_{\mathrm{JSD}}(p,q) = \tfrac{1}{2}D_{\mathrm{KL}}(p\,\|\,m) + \tfrac{1}{2}D_{\mathrm{KL}}(q\,\|\,m).
\label{eq:jsd}
\end{equation}

\section{Compressed Hindsight for Multi-Turn Self-Distillation}
\label{sec:problem}

On-policy self-distillation trains a policy on its own outputs while
conditioning a self-teacher on privileged information
\citep{hubotter2026reinforcement,zhao2026self,zhang2026learning}.
Most existing methods are developed for single-turn tasks. In a multi-turn
agent, however, the teacher must provide supervision for each intermediate
decision, and the choice of privileged context becomes nontrivial.

\subsection{\demosd degrades}
\label{sec:problem:degradation}

A straightforward extension is to give the teacher a successful trajectory
$\tau^{*}$ of the same task and use the teacher to re-score every student action
$y_t$. The complete demonstration provides rich hindsight about how the task
can be solved, and therefore provides a stronger source of privileged
information.

On \taur{}, the \demosd teacher reduces test success from $27.2$ to $22.5$,
below the untrained model (Table~\ref{tab:main}; Figure~\ref{fig:curves}
shows the training curve). In contrast, the \envsd teacher, which sees
only the immediate environment feedback, produces a modest improvement.
We attribute the degradation to what the demonstration exposes. At turn $t$,
the student chooses $y_t$ from the interaction history $\hist$. The
demonstration additionally exposes the teacher to actions that occur after turn $t$, including actions
from a different trajectory. The resulting teacher distribution can therefore
depend on how the demonstration proceeds after the current decision, rather
than only on evidence relevant to evaluating $y_t$.

Figure~\ref{fig:curves} suggests that more privileged
information is not necessarily better in multi-turn self-distillation. The
teacher context must provide useful hindsight without introducing information
poorly matched to the decision being supervised.

\subsection{Informative and locally aligned privileged context}
\label{sec:problem:axes}

The failure above highlights two properties of privileged context that matter
for multi-turn self-distillation:

\noindent\textbf{Informative.}
The teacher should receive enough evidence to identify why an action is
suboptimal. Immediate environment feedback is naturally associated with the
current action, but can be ambiguous. For example, an API error may reveal
that a call failed without identifying the earlier decision that caused the
failure.

\noindent\textbf{Locally aligned.}
The privileged context used to supervise turn $t$ should remain focused on
the decision made at turn $t$. This does not prohibit using future evidence:
later observations and the final outcome can be useful for diagnosing an
earlier action. The distinction is that the teacher should not directly
condition on the raw future actions when re-scoring the student's
original decision.

Environment feedback
is locally aligned but may be insufficient to explain the failure. A complete
successful demonstration is highly informative but exposes the teacher to a
future action sequence that is not part of the student's decision context.
Compressed hindsight is designed to retain the useful evidence from the
completed interaction while expressing that evidence as a turn-level signal.

\paragraph{Compressed hindsight.}
These observations motivate \method{}. A reflector can inspect the completed
trajectory and use later evidence to diagnose an earlier decision, but the
self-teacher receives only the resulting turn-level diagnosis rather than the
raw future trajectory. In this way, trajectory-level hindsight is compressed
before being used as privileged context. The rest of this section gives the
construction and the training objective.

\method uses the self-distillation setup of
Eq.~\eqref{eq:selfopd} with a new privileged context $c$. The
construction has three stages (Figure~\ref{fig:pipeline}): the student rolls
out normally, a reflector compresses the finished trajectory into per-turn
diagnoses, and a self-teacher re-scores the original action tokens under
those diagnoses.

\begin{table*}[t]
\centering\footnotesize
\setlength{\tabcolsep}{4pt}
\newcommand{\cellstrut}{\vphantom{$^{\dagger}$}}
\newcommand{\sr}[1]{\makebox[3.8em][l]{\makebox[2.0em][r]{#1\cellstrut}\makebox[1.8em][l]{}}}
\newcommand{\srd}[1]{\makebox[3.8em][l]{\makebox[2.0em][r]{#1\cellstrut}\makebox[1.8em][l]{$^{\dagger}$}}}
\newcommand{\srpm}[2]{\makebox[3.8em][l]{\makebox[2.0em][r]{#1\cellstrut}\makebox[1.8em][l]{\,{\tiny$\pm$#2}}}}
\newcommand{\turn}[1]{\makebox[3.8em][l]{\makebox[2.0em][r]{#1\cellstrut}\makebox[1.8em][l]{}}}
\newcommand{\turnpm}[2]{\makebox[3.8em][l]{\makebox[2.0em][r]{#1\cellstrut}\makebox[1.8em][l]{\,{\tiny$\pm$#2}}}}
\newcommand{\srhead}[1]{\makebox[3.8em][l]{\makebox[2.0em][c]{#1}\makebox[1.8em][l]{}}}
\newcommand{\turnhead}[1]{\makebox[3.8em][l]{\makebox[2.0em][c]{#1}\makebox[1.8em][l]{}}}
\begin{tabular}{lcccccc}
\toprule
& \multicolumn{2}{c}{\taur} & \multicolumn{2}{c}{\taua\ (OOD)} & \multicolumn{2}{c}{\webshop} \\
\cmidrule(lr){2-3}\cmidrule(lr){4-5}\cmidrule(lr){6-7}
\textbf{Design} & \srhead{SR $\uparrow$} & \turnhead{Turns $\downarrow$}
 & \srhead{SR $\uparrow$} & \turnhead{Turns $\downarrow$}
 & \srhead{SR $\uparrow$} & \turnhead{Turns $\downarrow$} \\
\midrule
\multicolumn{7}{@{}>{\columncolor{tabgroup}[0pt][0pt]}l@{}}{\textbf{Qwen3-4B}} \\
Base & \srd{27.2} & \turn{10.5} & \srd{12.0} & \turn{13.1} & \srd{21.8} & \turn{8.6} \\
\envsd & \srd{30.4} & \turn{11.5} & \srd{12.5} & \turn{13.7} & \srd{37.5} & \turn{10.4} \\
\demosd & \srd{\underline{22.5}} & \turn{12.4} & \srd{\underline{8.5}} & \turn{13.9} & \srd{\underline{17.4}} & \turn{10.6} \\
\sfthint & \srd{29.1} & \turn{11.2} & \srd{15.0} & \turn{13.6} & \srd{46.7} & \turn{10.3} \\
\rlhint & \srd{31.4} & \turn{12.6} & \srd{17.5} & \turn{14.3} & \srd{60.1} & \turn{10.6} \\
GRPO & \srpm{33.2}{1.1} & \turnpm{13.4}{0.8} & \srpm{17.8}{1.4} & \turnpm{14.0}{0.9} & \srpm{65.0}{0.8} & \turnpm{\textbf{8.1}}{0.7} \\
\method (ours) & \srpm{\textbf{34.9}}{0.5} & \turnpm{\textbf{9.6}}{0.4} & \srpm{\textbf{19.6}}{0.8} & \turnpm{10.1}{0.6} & \srpm{\textbf{68.8}}{1.0} & \turnpm{\textbf{8.1}}{0.4} \\
\midrule
\multicolumn{7}{@{}>{\columncolor{tabgroup}[0pt][0pt]}l@{}}{\textbf{Qwen3-30B-A3B}} \\
Base & \srd{33.6} & \turn{11.9} & \srd{30.0} & \turn{12.3} & \srd{32.7} & \turn{8.5} \\
\envsd & \srd{34.5} & \turn{12.2} & \srd{32.0} & \turn{12.6} & \srd{44.3} & \turn{9.3} \\
\demosd & \srd{\underline{26.1}} & \turn{12.6} & \srd{\underline{26.0}} & \turn{13.4} & \srd{\underline{25.9}} & \turn{9.9} \\
GRPO & \srpm{47.9}{0.8} & \turnpm{10.8}{0.6} & \srpm{33.1}{1.0} & \turnpm{11.6}{0.6} & \srpm{76.3}{1.2} & \turnpm{7.3}{0.5} \\
\method (ours) & \srpm{\textbf{50.1}}{0.4} & \turnpm{\textbf{10.7}}{0.5} & \srpm{\textbf{35.8}}{0.3} & \turnpm{\textbf{11.2}}{0.6} & \srpm{\textbf{79.9}}{0.7} & \turnpm{\textbf{7.0}}{0.5} \\
\bottomrule
\end{tabular}
\caption{\textbf{Main results.} Success rate (SR) and average number of turns across three benchmarks and two model scales. Bold indicates the best result, and underlining indicates performance below Base. $^{\dagger}$ denotes single-seed results; all other entries report mean $\pm$ standard deviation over three seeds. \outsd, which uses only the final outcome $R(\tau)$ as teacher context, is reported in Table~\ref{tab:ablation}.}
\label{tab:main}
\end{table*}

\subsection{Hindsight compression}
\label{sec:method:reflect}

Once a rollout terminates, a reflector $\mathcal{R}$---by default the policy
itself under a reflection prompt (Appendix~\ref{app:reflector})---reads the whole trajectory and its outcome
and emits one diagnosis per turn:
\begin{equation}
h_{1:T} = \mathcal{R}\big(\tau, R(\tau)\big), \qquad h_t = (d_t, \hat{a}_t),
\label{eq:reflect}
\end{equation}
where $d_t$ is a natural-language diagnosis of the decision at turn $t$ and
$\hat{a}_t$ an optional corrected action. When no actionable signal exists for
a turn, $\mathcal{R}$ emits $h_t = \varnothing$. We let $I(\tau)$ denote the
turns whose diagnosis parses and is non-empty; only these are scored.

The asymmetry between what $\mathcal{R}$ reads and what the teacher receives
is the mechanism of the method, not an implementation detail.
$\mathcal{R}$ sees everything, including $\tau_{>t}$ and $R(\tau)$, because
diagnosing turn $t$ requires knowing what happened afterwards. What
it passes on is a short, turn-indexed string. The teacher therefore inherits
the \emph{conclusion} of a backward pass over the trajectory without
inheriting the trajectory: it learns that an identity was fabricated at turn
$t$, not which tool call turn $t{+}3$ eventually used. The diagnosis is
therefore both informative and locally aligned, whereas a raw demonstration
is only informative (\S\ref{sec:problem:degradation}).

\subsection{Turn-level privileged self-teacher}
\label{sec:method:teacher}

For each $t\in I(\tau)$ we build the hindsight-augmented context
\begin{equation}
\phist = P\big(\hist,\, o_t,\, R(\tau),\, h_t\big),
\label{eq:context}
\end{equation}
with $P$ a fixed template (Appendix~\ref{app:prompt}). Each scored turn gets
its own prompt and its own forward pass; there is no shared context in which
a later turn could see the hindsight given for an earlier one. Relative to
the student's $\hist$, the teacher gains exactly three things: the immediate
observation $o_t$, the scalar outcome $R(\tau)$, and the diagnosis $h_t$. It
gains no action from any turn after $t$. Of the three, $h_t$ carries the bulk
of the signal and the other two are complementary: dropping $o_t$ and
$R(\tau)$ costs $1.0$ point, whereas a teacher with $o_t$ alone is $4.5$ points
below the full context (\S\ref{sec:exp:why}). We keep them because they are free, both are
already on hand when the rollout ends, and we do not claim them as the source
of the improvement.

The token-level distributions are
\begin{align}
p^{\text{stu}}_{t,k} &= \pi_\theta(\cdot\mid \hist,\, y_{t,<k}), \nonumber\\
p^{\text{tea}}_{t,k} &= \mathrm{sg}\big[\pi_{\bar\theta}(\cdot\mid \phist,\, y_{t,<k})\big],
\label{eq:dists}
\end{align}
and the objective averages the per-turn divergence over scored turns:
\begin{align}
\mathcal{L}_{\method}(\theta) &= \mathbb{E}_{\tau\sim\pi_\theta}
\Bigg[\frac{1}{|I(\tau)|}\sum_{t\in I(\tau)} \ell_t\Bigg], \nonumber\\
\ell_t &= \frac{1}{|y_t|}\sum_{k=1}^{|y_t|}
  D_{\mathrm{JSD}}\big(p^{\text{stu}}_{t,k},\, p^{\text{tea}}_{t,k}\big).
\label{eq:loss}
\end{align}
Both distributions are evaluated on the tokens the student actually produced.
The diagnosis $h_t$ and the corrected action $\hat{a}_t$ are never themselves
regression targets; they enter only through the teacher's conditioning. A
wrong diagnosis consequently costs a bounded, symmetric divergence on a
single turn rather than a copied mistake---which is precisely what separates
\method from supervising on $\hat{a}_t$ directly, a variant that loses $5.8$
points (\S\ref{sec:exp:main}).

\paragraph{Learning without reward contrast.}
Unlike GRPO, \method does not require reward variation across multiple rollouts.
The objective in Eq.~\eqref{eq:loss} remains defined whenever a turn receives a non-empty
diagnosis, including failed trajectories and the single-rollout setting
$\G=1$. \S\ref{sec:exp:sparse} evaluates these regimes empirically. Algorithm~\ref{alg:hint} summarizes one training step.

\section{Experiments}
\label{sec:exp}

\subsection{Setup}
\label{sec:exp:setup}

\paragraph{Benchmarks.}
We use six benchmarks, three for multi-turn agents and three for general
capability.
\begin{enumerate}[leftmargin=*,itemsep=1pt,topsep=2pt,parsep=0pt]
\item \textbf{\taur} \citep{yao2024tau}: a customer-service agent handles
order returns, exchanges and modifications for an LLM-simulated user via
tools under a domain policy. Success is judged by the final database state. We use $500$ training and $115$ test tasks.
\item \textbf{\taua} \citep{yao2024tau}: the airline domain of the same
benchmark, with flight bookings, changes and cancellations. Its $50$ test
tasks evaluate \taur-trained policies out of distribution.
\item \textbf{\webshop} \citep{yao2022webshop}: a simulated shopping website
on which the agent searches for and buys a product matching an instruction,
scored by a product-matching reward. We use $6{,}219$ training and $691$ test
tasks.
\item \textbf{MMLU} \citep{hendrycks2020measuring} and \textbf{MMLU-Pro}
\citep{wang2024mmlu}: multiple-choice questions over many subjects;
MMLU-Pro is harder, with ten options per question.
\item \textbf{IFEval} \citep{zhou2023instruction}: prompts with verifiable
format constraints, scored per prompt and per instruction, strict or loose.
\end{enumerate}

\paragraph{Models.}
We use Qwen3-4B-Instruct-2507, a dense model, and Qwen3-30B-A3B-Instruct-2507,
a mixture-of-experts model with $30$B total and $3$B active parameters
\citep{yang2025qwen3}. Both act through a ReAct-style interface in which each
turn is a reasoning span followed by one tool call \citep{yao2022react}. The
reflector is the policy itself; the teacher is its EMA copy with update rate
$0.001$.

\paragraph{Training.}
Each method trains on-policy for one epoch, with $16$
tasks per batch, at most $512$ generated tokens per turn and $8{,}192$ prompt
tokens. \method samples $\G=1$ rollout per task and GRPO samples $\G=8$.
\method minimizes the generalized JSD with $\alpha=0.5$ over the top-$100$
student tokens, with clipped importance weights ($\epsilon_{\min}=0.4$,
$\epsilon_{\max}=10$) and learning rate $2\times10^{-7}$. GRPO uses
$2\times10^{-6}$, the best of four values on the \taur validation split. The
4B model trains on 2$\times$A100 and the 30B model on 8$\times$H200 GPUs. See Appendix~\ref{app:impl} for details.

\paragraph{Evaluation.}
We report the mean@4 success rate (SR) and average turns, over three seeds
for GRPO and \method and one seed otherwise.

\paragraph{Baselines.}
We compare with GRPO \citep{shao2024deepseekmath} and five baselines that differ from \method in the teacher's context or in how
hindsight is used. The three self-distillation baselines keep
Eq.~\eqref{eq:loss} and replace the context of Eq.~\eqref{eq:context}.
\envsd gives the teacher the observation $o_t$, the execution feedback used by
SDPO \citep{hubotter2026reinforcement}. \demosd gives it a successful
trajectory $\tau^{*}$ of the same task, the reference-solution context of
prior self-distillation \citep{hubotter2026reinforcement,zhao2026self}.
\outsd gives it only the outcome $R(\tau)$, as in self-distillation from
binary rewards \citep{he2026self}. The other two consume the same reflections
as \method. \sfthint imitates the corrected action $\hat a_t$, as in training
on verbal feedback or hindsight-relabeled data
\citep{stephan2024rlvf,zhang2023wisdom}. \rlhint turns the teacher--student
divergence into a turn-level process reward for RL \citep{lightman2024let}.
Base is the untrained instruction-tuned model.

\subsection{Main results}
\label{sec:exp:main}

Tables~\ref{tab:main} and~\ref{tab:costmain} support four conclusions.

\paragraph{The teacher's context, not dense supervision, causes the degradation.}
\demosd is the only design that falls below Base, on every benchmark and at
both scales. \envsd optimizes the same objective with a less informative
context and improves over Base. Dense self-distillation is therefore not
harmful in itself. The harm comes from a context with later actions to copy.

\paragraph{Compressed hindsight is the strongest design at both scales.}
\method has the highest success rate in every benchmark--scale cell,
including against GRPO. \method is also the only self-distillation design
that shortens interactions relative to Base; every other design lengthens
them. Its gains therefore do not come from taking more turns.

\paragraph{\method is better at half the training cost.}
On \taur, \method reaches higher success than GRPO with about half the
GPU-hours (Table~\ref{tab:costmain}). Each trajectory costs more, with a
reflection call and a teacher forward pass per scored turn, but one rollout
per task instead of eight more than offsets this.

\begin{table}[!t]
\centering\small
\setlength{\tabcolsep}{4pt}
\begin{tabular}{@{}lcc@{}}
\toprule
 & GRPO & \method \\
\midrule
Rollouts per task & 8 & 1 \\
Generated tokens per step & 420.2K & 62.5K \\
Time per step (s) & 557.3 & 282.6 \\
Total GPU-hours & 9.60 & \textbf{4.87} \\
\taur SR & 33.2 & \textbf{34.9} \\
\bottomrule
\end{tabular}
\caption{\textbf{Training cost on \taur{} (4B).}}
\label{tab:costmain}
\end{table}

\paragraph{How hindsight is consumed matters as much as what it contains.}
\sfthint and \rlhint use the same reflections as \method yet fall short of it
on every benchmark. \sfthint imitates the corrected action $\hat a_t$, so each
wrong reflection becomes a target, and about a third are incorrect or vacuous
(\S\ref{sec:analysis:reflection}). \rlhint collapses the token-level
divergence into one reward per turn, losing where the error lies. \method avoids both: the reflection only conditions a per-token
teacher.

\section{Analysis}
\label{sec:analysis}

We study four questions. \S\ref{sec:exp:why} ablates the teacher's context and
locates its supervision. \S\ref{sec:exp:sparse} tests learning
under scarce reward contrast, \S\ref{sec:analysis:reflection} measures hint
quality, and \S\ref{sec:analysis:general} checks general capability.

\subsection{Why compressed hindsight works}
\label{sec:exp:why}

We train one Qwen3-4B seed per teacher context, otherwise as in
\S\ref{sec:exp:setup}. To locate supervision, we split the teacher--student
divergence on \taur wrong-tool turns over tool-name, argument and reasoning
tokens.

\begin{table}[!h]
\centering\small
\setlength{\tabcolsep}{3.5pt}
\begin{tabular}{cccccc}
\toprule
$o_t$ & $R(\tau)$ & $h_t$ & Retail & Airline & WebShop \\
\midrule
\checkmark & & & 30.4 & 12.5 & 37.5 \\
& \checkmark & & 28.2 & 13.0 & 33.9 \\
& & \checkmark & 33.9 & 18.5 & 63.1 \\
\checkmark & & \checkmark & 34.0 & 19.0 & 64.6 \\
\checkmark & \checkmark & \checkmark & \textbf{34.9} & \textbf{19.6} & \textbf{68.8} \\
\midrule
\multicolumn{6}{l}{\emph{Divergence mass over all wrong-tool turns}} \\
\multicolumn{3}{l}{Teacher} & Tool & Args & Reason. \\
\multicolumn{3}{l}{\envsd} & 2.2\% & 39.8\% & 58.0\% \\
\multicolumn{3}{l}{\method} & \textbf{37.6\%} & 9.1\% & 53.3\% \\
\bottomrule
\end{tabular}
\caption{\textbf{Teacher-context ablation (top) and share of the divergence on
wrong-tool turns (bottom)}, 4B.}
\label{tab:ablation}
\end{table}

\paragraph{\method supervises the cause of an error, not its symptom.}
Figure~\ref{fig:tokloss} shows one wrong-tool turn. Instead of asking for the
user's name, the agent fabricates one and calls
\texttt{find\_user\_id\_by\_name\_zip}; the environment returns only
\texttt{user not found}. The \envsd teacher reads the error message and
concentrates its loss on the rejected name. The \method teacher, given the
diagnosis, concentrates it on the tool name instead. The bottom of
Table~\ref{tab:ablation} shows that this case is typical: across all
wrong-tool turns, \method moves the divergence from the arguments to the tool
name. Environment feedback reports where an error surfaces, a diagnosis
reports where it was made, and the teacher's divergence follows the one it is
given.

\paragraph{The diagnosis carries most of the gain.}
The top of Table~\ref{tab:ablation} ablates the three parts of $\phist$. The
diagnosis alone recovers most of the full context's gain over Base, while the
observation or the outcome alone recovers less than half of it. \demosd marks
the other extreme: hindsight given raw, as a full demonstration, lowers
success below Base.

\begin{figure}[t]
\centering
\includegraphics[width=\linewidth]{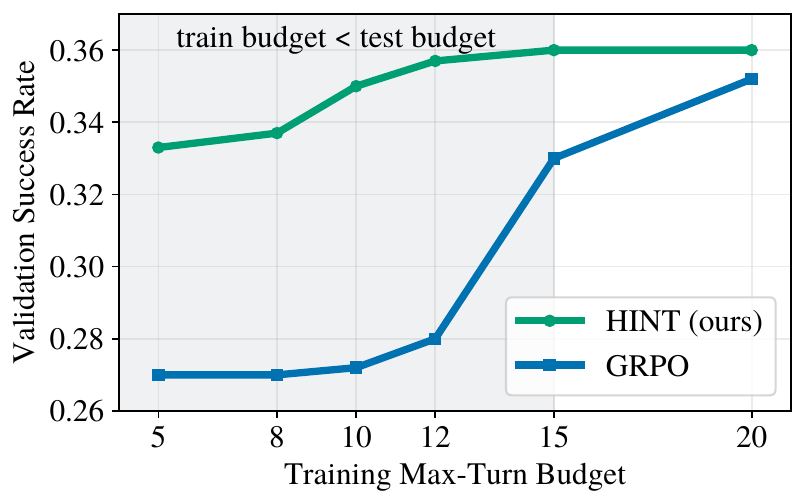}
\caption{\textbf{Validation success against the training turn budget on
\taur.} Shading marks training budgets below the test budget of $15$, where
successful rollouts are rare and reward contrast collapses.}
\label{fig:budget}
\end{figure}

\subsection{Learning beyond sparse outcome rewards}
\label{sec:exp:sparse}

\begin{figure}[!h]
\centering
\includegraphics[width=\linewidth]{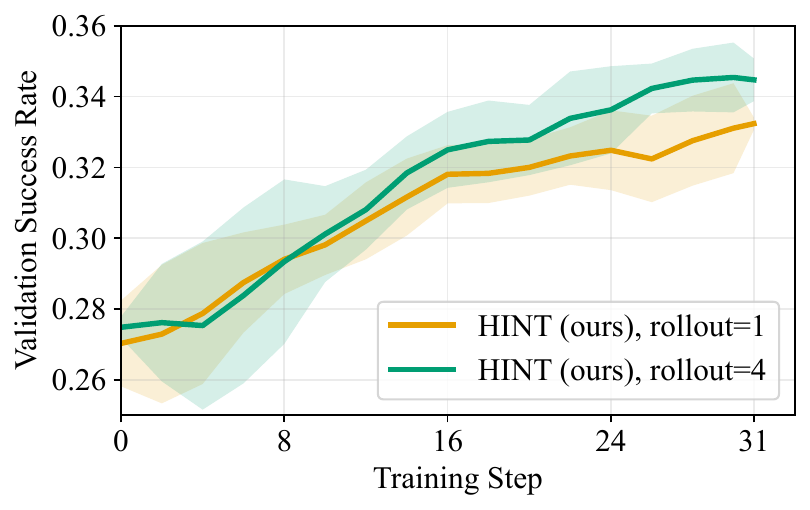}
\caption{\textbf{Validation success of \method with $\G=1$ and $\G=4$ on
\taur}, shaded over three seeds.}
\label{fig:rollout}
\end{figure}

\method draws no supervision from reward contrast, so it remains usable in
two regimes where GRPO runs out of signal. Both experiments train Qwen3-4B on
\taur. For Figure~\ref{fig:budget}, the training turn budget ranges from $5$
to $20$, and every checkpoint is evaluated with the test budget of $15$. For
Figure~\ref{fig:rollout}, \method trains with $\G=1$ and $\G=4$ over three
seeds. Figure~\ref{fig:budget} compares
\method and GRPO as the training turn budget tightens: most rollouts then fail before
reaching a terminal success, group-relative advantages lose their contrast,
and GRPO flattens while \method keeps improving. At $\G=1$,
GRPO's advantage is identically zero, while \method recovers most of its
$\G=4$ performance at a quarter of the interaction cost
(Figure~\ref{fig:rollout}).

\subsection{Reflection quality}
\label{sec:analysis:reflection}

\paragraph{Protocol.}
We sample $75$ self-reflector turn-level hints from successful, failed and
budget-exhausted \taur training rollouts. The annotator sees $\hist$, $y_t$,
$o_t$ and $h_t$. One author labelled all $75$ hints.

\paragraph{Rubric.}
A \emph{correct} hint identifies a real execution error or confirms a sound
action, with any correction consistent with the tool schema and task state. A
\emph{partially correct} hint finds a real issue but misattributes its cause
or turn, or proposes a suboptimal repair. An \emph{incorrect} hint is
factually wrong, violates the schema, or uses information unavailable at turn
$t$. A \emph{vacuous} hint carries no usable supervision, e.g.\ it marks a
failed trajectory as correct or names an error without a repair.

\begin{table}[!h]
\centering\small
\setlength{\tabcolsep}{4pt}
\begin{tabular}{lc}
\toprule
Label & Self-reflector \\
\midrule
Correct & 58.6\% \\
Partially correct & 9.3\% \\
Incorrect & 17.3\% \\
Vacuous / non-actionable & 14.7\% \\
\bottomrule
\end{tabular}
\caption{\textbf{Manual annotation of $75$ self-reflector hints.}}
\label{tab:reflection}
\end{table}

\paragraph{\method tolerates imperfect hints.}
The self-reflector writes incorrect or vacuous hints in about a third of cases
(Table~\ref{tab:reflection}). The teacher only conditions on a hint, so a wrong
hint perturbs the supervision of one turn instead of becoming a target.
Three failure modes recur. \emph{Hindsight leakage} cites information
available only in later turns, \emph{recency bias} blames the latest turn
rather than the earlier fork, and \emph{over-confident correction} proposes a
plausible but unsupported next action.

\paragraph{Parsing-based selection.}
Supervision is applied only to turns whose structured hint parses and is
non-empty, so a malformed hint drops its turn from the loss. Parse failures
affect $4.8\%$ of turns for the 4B model and $3.2\%$ for the 30B model.

\subsection{General capability preservation}
\label{sec:analysis:general}

Training on narrow agent data can affect general capabilities, especially with
fixed off-policy data. We therefore evaluate Base and the final \taur
checkpoints of GRPO and \method on MMLU and MMLU-Pro, which test knowledge and
reasoning, and on IFEval, which tests instruction following.

\begin{table}[!h]
\centering\small
\setlength{\tabcolsep}{3.2pt}
\resizebox{\linewidth}{!}{%
\begin{tabular}{@{}lcccccc@{}}
\toprule
\multirow{2}{*}{Method} & \multirow{2}{*}{MMLU} & \multirow{2}{*}{MMLU-Pro} & \multicolumn{2}{c}{IFEval instruction} & \multicolumn{2}{c}{IFEval prompt} \\
\cmidrule(lr){4-5}\cmidrule(l){6-7}
& & & Loose & Strict & Loose & Strict \\
\midrule
Base & 59.60 & 34.38 & 89.93 & 88.13 & 85.40 & 82.62 \\
GRPO & 59.61 & 34.25 & 89.80 & 88.60 & 86.70 & 82.19 \\
\method & 59.54 & 34.35 & 89.68 & 88.13 & 86.32 & 82.80 \\
\bottomrule
\end{tabular}}
\caption{\textbf{General capability preservation.}}
\label{tab:general}
\end{table}

Differences from Base are small and go in both directions (Table~\ref{tab:general}), so we read them as
capability preservation rather than improvement. The result matches findings
that on-policy updates preserve capabilities better than training on fixed data
\citep{chen2025retaining,shenfeld2025rl} and that self-distillation supports
continual learning \citep{shenfeld2026selfdistillationenablescontinuallearning}.

\section{Related Work}
\label{sec:related}

\paragraph{On-policy self-distillation with privileged information.}
On-policy distillation trains a student on its own sampled outputs rather
than fixed teacher-generated data, reducing the train--deploy mismatch of
off-policy distillation
\citep{agarwal2024policy,lu2025onpolicydistillation}.
Self-distillation further removes the external teacher by evaluating the same
model under privileged information unavailable to the student
\citep{snell2022learning}.
SDPO conditions the self-teacher on execution feedback or a successful
solution and distills the resulting distribution back to the original policy
\citep{hubotter2026reinforcement}.
Other recent methods provide reference solutions, hints, or self-generated
revisions as privileged context for the teacher
\citep{zhao2026self,ye2026policy,penaloza2026privileged,he2026self}.
ReSD is especially related in combining reflection with self-distillation
\citep{zhang2026learning}.
These methods are developed primarily for single-turn outputs, whereas we
study how privileged information should be constructed when a trajectory
contains multiple decisions that must be supervised separately.

\paragraph{Reflection and hindsight as training signals.}
Reflection methods such as Reflexion and Self-Refine use natural-language
critique to revise an output or guide subsequent attempts at inference time
\citep{shinn2023reflexion,madaan2023self,saunders2022self}.
Reflection can also be converted into training supervision:
TextGrad treats verbal feedback as a gradient-like signal for prompt
optimization \citep{yuksekgonul2024textgrad}, while RLVF and related methods
use verbal feedback or pre/post-reflection outputs to construct preference or
training data \citep{stephan2024rlvf}.
Hindsight methods instead relabel completed trajectories using information
available only after the outcome, as in hindsight experience replay and its
language-model variants
\citep{Andrychowicz2017HindsightER,liu2024chain,zhang2023wisdom}.
\method also uses post-hoc information, but the reflected diagnosis is not
directly imitated or optimized; it is used only to condition a
self-teacher that re-scores the student's original action.

\paragraph{Credit assignment in multi-turn agent learning.}
Multi-turn agents are commonly trained with terminal task rewards on
benchmarks such as WebShop, $\tau$-Bench, and AppWorld
\citep{yao2022webshop,yao2024tau,trivedi2024appworld}.
When intermediate supervision is available, prior work improves credit
assignment through reward shaping, hierarchical values, structured
advantages, or process reward models
\citep{ahmad2024accelerating,zhou2024archer,zhang2026reasoning,
lightman2024let}.
GRPO can also induce intermediate credit through group-relative outcome
advantages \citep{sullivan2026grpo}.
Our setting instead assumes only terminal automatic feedback and obtains
turn-level supervision by conditioning a self-teacher on compressed hindsight.
\section{Conclusion}
\label{sec:conclusion}

Carrying single-turn self-distillation into multi-turn agents fails in an
informative way: handing the teacher a successful demonstration at every turn
is dense and informative, and drives the policy below where it started. The
reason is that privileged context must also be \emph{locally aligned} with
the decision being scored. \method combines dense token-level self-distillation
with compressed hindsight, giving localized supervision without handing the
teacher a future it can imitate. The resulting alignment is measurable: the
share of teacher--student divergence landing on the erroneous decision rises
from $2.2\%$ to $37.6\%$.

\section*{Limitations}
\method depends on the policy's ability to diagnose its own trajectories in
hindsight. It is best suited to errors recognizable after the fact, such as
malformed tool calls, fabricated arguments and repeated API calls; we do not
claim it helps when the model cannot diagnose the failure even in retrospect.

The method also inherits reflector errors. Roughly a third of self-generated
hints are incorrect or vacuous (\S\ref{sec:analysis:reflection}), and malformed
hints are dropped by parsing. The parse rate is high, $95.2\%$ for 4B and
$96.8\%$ for 30B, but this selection effect is not eliminated.

Finally, our statistical evidence is uneven. The degradation result and the
comparison against other self-distillation designs involve large effects, but
the 4B margin over GRPO on the two \taubench domains is small. \method also
costs more per trajectory than GRPO, despite using fewer rollouts.

\bibliography{custom}

\appendix

\section{Notation}
\label{app:notation}

\begin{table}[h]
\centering\footnotesize
\setlength{\tabcolsep}{4pt}
\begin{tabular}{ll}
\toprule
Symbol & Meaning \\
\midrule
$x$ & task instruction \\
$y_t$ & action at turn $t$ (reasoning + tool call) \\
$o_t$ & environment observation returned for $y_t$ \\
$\hist$ & history $(x,y_1,o_1,\dots,y_{t-1},o_{t-1})$ \\
$\tau$ & rollout $(x,y_1,o_1,\dots,y_T,o_T)$ \\
$\tau_{>t}$ & trajectory suffix after turn $t$ \\
$\tau^{*}$ & successful demonstration (another model) \\
$R(\tau)$ & terminal reward of the rollout \\
$T$ & turn budget \\
$\G$ & rollouts sampled per task instruction \\
$y_{t,k}$ & $k$-th token of $y_t$; $y_{t,<k}$ its prefix \\
$\mathcal{R}$ & reflector \\
$h_t=(d_t,\hat a_t)$ & turn-level diagnosis and corrected action \\
$I(\tau)$ & scored turns ($h_t$ parses, non-empty) \\
$\phist$ & teacher context $P(\hist,o_t,R(\tau),h_t)$ \\
$\pi_\theta,\pi_{\bar\theta}$ & student policy, EMA teacher copy \\
\bottomrule
\end{tabular}
\caption{Notation used throughout.}
\end{table}

\section{Training Algorithm}
\label{app:alg}

Algorithm~\ref{alg:hint} gives one training step. Its loss is
Eq.~\eqref{eq:loss} with each token's divergence weighted by a clipped
importance weight $w_{t,k}$ (Appendix~\ref{app:impl}).

\begin{algorithm}[ht]
\caption{One training step of \method. All rollouts are kept; only turns
with a non-empty diagnosis enter the loss.}
\label{alg:hint}
\begin{algorithmic}[1]
\Require student $\pi_\theta$, teacher $\pi_{\bar\theta}$, reflector $\mathcal{R}$
\LComment{Stage 1: one rollout per task ($\G=1$)}
\LComment{$y_t$: action; $o_t$: tool output; $R(\tau)$: reward}
\For{each task $x$ in a batch $\mathcal{B}$}
  \State $\tau=(x,y_1,o_1,\dots,y_T,o_T)\sim\pi_\theta$
  \LComment{Stage 2: reflector diagnoses each turn (App.~\ref{app:reflector})}
  \State $h_{1:T}\gets\mathcal{R}\big(\tau,R(\tau)\big)$
    \Comment{Eq.~\eqref{eq:reflect}}
  \State $I(\tau)\gets\{\,t : h_t\neq\varnothing\,\}$
  \LComment{Stage 3: score the student's own tokens}
  \For{each $t\in I(\tau)$}
    \State $\hist\gets(x,y_1,o_1,\dots,y_{t-1},o_{t-1})$
    \LComment{teacher: no action after turn $t$}
    \State $\phist\gets P\big(\hist,o_t,R(\tau),h_t\big)$
    \State $p^{\text{stu}}_{t,k}\gets\pi_\theta(\cdot\mid\hist,y_{t,<k})$
    \State $p^{\text{tea}}_{t,k}\gets\mathrm{sg}\big[\pi_{\bar\theta}(\cdot\mid\phist,y_{t,<k})\big]$
    \LComment{$w_{t,k}$: clipped importance weight }
    \State $\ell_t\gets\frac{1}{|y_t|}\sum_{k} w_{t,k}\,D_{\mathrm{JSD}}\big(p^{\text{stu}}_{t,k},p^{\text{tea}}_{t,k}\big)$
  \EndFor
\EndFor
\LComment{Stage 4: update (empty $I(\tau)$ adds 0 )}
\State $\mathcal{L}\gets\frac{1}{|\mathcal{B}|}\sum_{\tau}\frac{1}{\max(1,|I(\tau)|)}\sum_{t\in I(\tau)}\ell_t$
\State $\theta\gets\theta-\eta\,\nabla_\theta\mathcal{L}$
\State $\bar\theta\gets(1-\lambda)\,\bar\theta+\lambda\,\theta$
  \Comment{EMA teacher}
\end{algorithmic}
\end{algorithm}

\section{Implementation Details}
\label{app:impl}

\paragraph{Training.} All methods train for one epoch with a batch of 16
prompts; rollouts are generated on-policy rather than replayed from a buffer.
\method uses $\G=1$ rollout per prompt in all main experiments. We cap
generation at 512 tokens per assistant turn and the prompt at 8192 tokens.
Validation reports mean@4 over four samples per task.

\paragraph{Objective.} \method uses the generalized Jensen--Shannon
divergence with $\alpha=0.5$. The divergence is computed over the student's
top-100 tokens at each position, with a tail bucket for the remaining mass, and applied only to assistant-generated tokens---user
turns, tool observations and environment feedback are masked out. The teacher
is an EMA copy with update rate $0.001$. Clipped importance weighting uses
$\epsilon_{\min}=0.4$ and $\epsilon_{\max}=10$.

\paragraph{Hardware.} Qwen3-4B-Instruct trains on 2$\times$A100 (80GB) and
Qwen3-30B-A3B-Instruct on 8$\times$H200. Every baseline uses the same
configuration at the same scale, so the cost comparison in
Table~\ref{tab:costmain} is at matched hardware.

\paragraph{Learning rates.} We tune GRPO's learning rate separately on the
\taur validation split rather than forcing a shared value, which favours the
baseline: $2\times10^{-7}\!\to\!32.4$, $5\times10^{-7}\!\to\!33.0$,
$2\times10^{-6}\!\to\!33.2$, $5\times10^{-6}\!\to\!31.9$, so GRPO uses
$2\times10^{-6}$. \method uses $2\times10^{-7}$, which is stable because its
token-level supervision is far denser than a terminal reward; the two
objectives differ in gradient density and scale, and matching learning rates
across them would not be the fairer choice.

\paragraph{Datasets.} \taur has 500 training and 115 test tasks, \webshop
6219 and 691, and \taua 50 test tasks for out-of-distribution evaluation only
\citep{yao2024tau,yao2022webshop}. \taubench checks database post-conditions
and \webshop scores product matching.

\section{Teacher Prompt}
\label{app:prompt}

Figure~\ref{fig:prompt} shows the template $P$ of Eq.~\eqref{eq:context}.
The reflector prompt that produces $h_t$ is in Appendix~\ref{app:reflector}.

\begin{figure}[h]
\definecolor{hintGreen}{HTML}{009E73}
\definecolor{hintGreenSoft}{HTML}{D3EFE6}
\definecolor{actYellow}{HTML}{F6E7B8}
\definecolor{slotInk}{HTML}{2C3E55}

\newcommand{\role}[1]{\colorbox{black!8}{\sffamily\bfseries\scriptsize #1}}
\newcommand{\slot}[2]{\textcolor{slotInk}{\{\{#1\}\}}%
  \hfill{\color{black!45}\normalfont\scriptsize$#2$}}

\begin{tcolorbox}[enhanced, colback=white, colframe=black!15,
  boxrule=0.5pt, arc=2pt, left=6pt, right=6pt, top=4pt, bottom=4pt,
  title={\small\sffamily\bfseries Teacher context $\widetilde{H}_t$},
  coltitle=black, colbacktitle=black!5,
  fontupper=\ttfamily\footnotesize]
\role{User}\\
\slot{original\_task\_prompt}{x}\\
\slot{interaction\_history}{y_{<t},\,o_{<t}}\\[2pt]
Feedback on your earlier attempt:\\
Outcome: \slot{trajectory\_success}{R(\tau)}\\
Tool observation: \slot{environment\_output}{o_t}\\
\colorbox{hintGreenSoft}{\parbox{\dimexpr\linewidth-2\fboxsep}{%
  Reflection: \slot{reflection\_hint}{h_t}}}\\[2pt]
Use the hint when evaluating the original action for this turn.\\[3pt]
\role{Assistant}\\
\colorbox{actYellow}{\parbox{\dimexpr\linewidth-2\fboxsep}{%
  \slot{original\_response}{y_t\ \text{(scored tokens)}}}}
\end{tcolorbox}
\caption{The teacher prompt template $P$ of Eq.~\eqref{eq:context}. The
student is scored on the highlighted original action tokens; the hindsight
diagnosis $h_t$ enters only as context and is never itself a target.}
\label{fig:prompt}
\end{figure}

\section{Reflector Prompt}
\label{app:reflector}

Figure~\ref{fig:reflector} shows the reflection prompt of
Eq.~\eqref{eq:reflect}. After each rollout, $\mathcal{R}$ is called once on
the task, tool schema, full trajectory $\tau$ and outcome $R(\tau)$, and returns one structured hint per assistant turn. The
\texttt{diagnosis} field is $d_t$, and \texttt{next\_action\_target}, when
present, is the corrected action $\hat a_t$. In all main experiments
$\mathcal{R}$ is the policy itself.

\begin{figure*}[t]
\begin{tcolorbox}[enhanced, colback=white, colframe=black!15,
  boxrule=0.5pt, arc=2pt, left=6pt, right=6pt, top=4pt, bottom=4pt,
  title={\small\sffamily\bfseries Reflection prompt},
  coltitle=black, colbacktitle=black!5, fontupper=\footnotesize]
\textbf{Role.} You are an observation-grounded hindsight critic for a
multi-turn tool-use agent. Given the original task, tool schema, full
trajectory, and final reward, produce one structured JSON hint for each
assistant turn.

\smallskip\textbf{Rules.}
\begin{enumerate}\setlength{\itemsep}{0pt}\setlength{\parsep}{0pt}\setlength{\topsep}{2pt}
\item Reflect on every assistant turn, including correct turns.
\item Ground each diagnosis in the current action and its next environment
  observation. Use later trajectory evidence only to identify delayed
  consequences; do not expose future actions or observations verbatim.
\item Distinguish decision errors from tool-execution errors.
\item Fill \texttt{next\_action\_target} only when the corrected action is
  unambiguous; otherwise set it to \texttt{null} with \texttt{"medium"} or
  \texttt{"low"} confidence.
\item Avoid redundant actions, repeated confirmations, and re-asking already
  confirmed information.
\item Output only valid JSON, with no markdown or extra text.
\end{enumerate}

\smallskip\textbf{Output format.} Return a JSON object with one key per
assistant turn: \texttt{"turn1"}, \texttt{"turn2"}, \dots, \texttt{"turnT"}.
\begin{verbatim}
{ "turn1": { "diagnosis": "<brief causal diagnosis>",
             "confidence": "low",
             "next_action_target": null },
  "turn2": { "diagnosis": "<brief causal diagnosis>",
             "confidence": "high",
             "next_action_target": {
               "thought": "<one-line local plan>",
               "action": { "name": "<tool_name_or_respond>",
                           "arguments": {"<arg_name>": "<arg_value>"} } } } }
\end{verbatim}
\end{tcolorbox}
\caption{The reflection prompt of Eq.~\eqref{eq:reflect}. The reflector
returns one structured hint per assistant turn; rule~2 keeps each hint
grounded in the current turn and forbids copying later actions into it.}
\label{fig:reflector}
\end{figure*}

\end{document}